\documentclass[runningheads]{llncs}
\usepackage[T1]{fontenc}
\usepackage{graphicx}
\usepackage{booktabs}
\usepackage[misc]{ifsym}
\newcommand{\corr}{(\Letter)}
\usepackage{mwe}
\usepackage[usenames,dvipsnames]{xcolor}
\usepackage{color, colortbl}
\definecolor{Gray}{gray}{0.9}
\definecolor{LightCyan}{rgb}{0.88,1,1}
\usepackage{amsmath,amssymb,amsfonts}
\usepackage{algorithmic}
\usepackage{textcomp}
\usepackage{subcaption}
\usepackage{array, makecell}
\usepackage{enumitem}

\begin{document}

\title{Adaptive Knowledge Distillation for Classification of Hand Images using Explainable Vision Transformers}

\titlerunning{Adaptive Knowledge Distillation for Classification using Explainable ViTs}



\author{Thanh Thi Nguyen\inst{1} \corr \and
Campbell Wilson\inst{1} \and
Janis Dalins\inst{2}}


\authorrunning{T.T. Nguyen et al.}


\institute{AiLECS Lab, Monash University, Melbourne VIC 3800, Australia \email{\{thanh.nguyen9,campbell.wilson\}@monash.edu}
\and
AiLECS Lab, Australian Federal Police, Melbourne VIC 3800, Australia \email{janis.dalins@afp.gov.au}
}

\maketitle              

\begin{abstract}

Assessing the forensic value of hand images involves the use of unique features and patterns present in an individual's hand. The human hand has distinct characteristics, such as the pattern of veins, fingerprints, and the geometry of the hand itself. This paper investigates the use of vision transformers (ViTs) for classification of hand images. We use explainability tools to explore the internal representations of ViTs and assess their impact on the model outputs. Utilizing the internal understanding of ViTs, we introduce distillation methods that allow a student model to adaptively extract knowledge from a teacher model while learning on data of a different domain to prevent catastrophic forgetting. Two publicly available hand image datasets are used to conduct a series of experiments to evaluate performance of the ViTs and our proposed adaptive distillation methods. The experimental results demonstrate that ViT models significantly outperform traditional machine learning methods and the internal states of ViTs are useful for explaining the model outputs in the classification task. By averting catastrophic forgetting, our distillation methods achieve excellent performance on data from both source and target domains, particularly when these two domains exhibit significant dissimilarity. The proposed approaches therefore can be developed and implemented effectively for real-world applications such as access control, identity verification, and authentication systems.

\end{abstract}

\section{Introduction}
\label{sec_int}

Common biometric features encompass fingerprints, hand geometry, iris patterns, facial features, voice patterns, and even typing patterns or gait \cite{dargan2020comprehensive}. The choice of which feature to use depends on factors such as security requirements, convenience, and the specific use case. Hand images have become a popular biometric feature because the physical characteristics of an individual's hand are unique and can be reliably measured and analyzed \cite{aftab2021hand}. Hand features are relatively stable over time, meaning they do not change significantly with age. This makes hand-based biometrics a reliable method \cite{jia2021survey}. Furthermore, capturing hand images is non-intrusive and can be done using various devices such as cameras or scanners. Compared to some other biometric modalities, e.g., fingerprints, using hand images is often considered more hygienic as it does not involve direct contact with a scanning surface \cite{okereafor2020fingerprint}.

Effective yet transparent and explainable biometric systems are valuable, particularly in the context of law enforcement where decisions can have significant consequences. Machine learning techniques have found extensive use in biometric applications involving image data. The recent advancements in deep learning, particularly the emergence of vision transformer (ViT)~\cite{dosovitskiy2021an}, have substantially elevated the field of image processing. In this research, we explore the application of ViTs for biometric systems, specifically utilizing hand images as biometric features. To unveil the workings of the ViT model and provide a glimpse into its internal mechanisms, we employ the deep feature factorization \cite{collins2018deep} and gradient-weighted class activation mapping (Grad-CAM)~\cite{selvaraju2020grad} tools to generate insightful visualizations, revealing what the models perceive within images.

More significantly, we endeavor to tackle the issue where a model trained on data from a specific domain exhibits suboptimal performance when applied to data from a distinct domain. For example, a ViT model trained on palm hand images does not exhibit satisfactory performance when presented with dorsal hand images, and vice versa. A common remedy for this is to fine-tune the model using data from the new domain. Nevertheless, this approach gives rise to the issue of catastrophic forgetting, where the model tends to erase knowledge acquired from the previous (source) domain \cite{binici2022preventing}. Several studies have suggested incorporating a portion of data from the source domain in the fine-tuning process to help the model retain previous knowledge \cite{lopez2017gradient,chaudhry2021using,davari2022probing}. However, in the realm of biometric systems, particularly in law enforcement applications, there are instances where access to source domain data is restricted due to privacy concerns and/or lack of data consents.
We therefore propose domain adaptation methods to facilitate the fine-tuning of ViT models on new domain data while retaining the knowledge gained from the source domain data.

In summary, this work makes a triple-fold contribution: 1) fine-tune 6 ViT models for classifying hand images and compare their performance with existing methods; 2) investigate various components of the black-box ViT models, assessing their influence on model outputs through explainability tools; and 3) propose adaptive distillation methods, enabling the ViT model to excel in a new domain while retaining knowledge acquired previously without access to source domain data. 

\section{Important Caveats and Application}
It is important to emphasize that no elements of this research are proposed for use in identifying or storing individuals’ sensitive or private data, including biometrics, and it does not involve the gathering of additional data, nor data outside of existing legal instruments available to police. This work's role is to merely rank imagery by likely value to a forensic examiner, in line with existing legally accepted processes and controls.

While we use the term \emph{classification} in line with other work in this field, we also note that this does not imply for us the identification of individuals. We use the term purely for consistency in comparison with other results. Furthermore, such classification does not meet forensic standards such that these techniques would ever be solely relied on for identification. The work in this paper is intended purely to assist in triage of images likely to be of value to qualified forensic examiners.

We further outline ethical considerations later in the paper. It is important to note that individual use-case scenarios of these technologies should always be subject to through assessment of the ethical ramifications, for example against frameworks such as \cite{ANZPAA}.

\section{Related Work}
\label{sec_rel_wks}

The structure of the hand, including the length and width of fingers, the shape of the palm, and the configuration of veins, forms a set of distinctive features that vary from person to person. Hardalac et al. \cite{hardalac2020novel} incorporated various feature extraction methods such as ripplet-I transform (an extended version of curvelet transform), discrete cosine transform, discrete wavelet transform, contourlet transform, PCA, and LBP for palm print image processing. A traditional artificial neural network, Euclidean distance, SVM, and CNN are used as classifiers. The most effective contributors to palm print verification and identification are found to be the LBP features and the Euclidean distance classifier.

Likewise, a rapid palmprint recognition system, named Simplified PalmNet-Gabor, which leverages the PalmNet was designed in \cite{trabelsi2022efficient}. The Log-Gabor filters are used to modify the pixel luminance of palm print images. Fisher score and ReliefF are then employed for feature selection while the whitening PCA method is used for dimensionality reduction, aiming to decrease computational expenses and enhance accuracy with SVM and k-nearest neighbour classifiers. In another study, ViT models were employed in \cite{garcia2023vision} for vein biometric recognition applications. ViTs demonstrate better performance compared with other methods across different datasets.

On the other hand, Afifi \cite{afifi201911k} introduced a large dataset of 11k hand images with useful metadata and used CNN and LBP as feature extraction methods and SVM as a classifier for gender recognition and biometric identification purposes. Four different CNN architectures including AlexNet \cite{krizhevsky2012imagenet}, VGGNet (16- and 19-layer variants)~\cite{simonyan2015very}, and GoogleNet \cite{szegedy2015going} are applied to extract deep features. Hand-crafted feature extraction methods are also used such as scale invariant feature transform (SIFT), color-invariant SIFT \cite{charfi2014novel}, and \textit{rg}SIFT~\cite{van2009evaluating}. The distinctive features on the dorsal side are discovered to be effective, comparable to, if not superior to, those present in palmar side images of human hands.

The work closest to our study is presented in \cite{zhao2021continual}, where they introduced a framework for continual representation learning by employing knowledge distillation with neighborhood selection and consistency relaxation modules for face recognition and person re-identification applications. That problem is framed as a class-incremental classification task \cite{kang2022class}. Their configuration restricts the model to accessing only the training data from the current learning step. Lacking access to previous classes, the model progressively loses the knowledge acquired during earlier learning steps. The training process involves several steps, with each step dedicated to training on images from newly introduced classes, where the visual appearance of distinct classes can be highly relevant. Their approach shares similarities with ours in that it prevents the model from accessing previous training data. However, our setting differs from theirs; we are not specifically addressing the class-incremental problem. Instead, our objective is to adapt the model to data from a new domain, which typically lacks visual relevance to data from the old domain. As an illustration, consider a model initially trained on palmar hand images and then adapted to handle dorsal hand images, or vice versa. Despite both domains capturing human hands, the features and appearances of the palmar and dorsal sides exhibit significant differences.

\section{ViT Models}
\label{sec_vit_models}
The ViT was first introduced in \cite{dosovitskiy2021an} by Google based on the transformer architecture proposed to address natural language processing tasks. We refer to this initial model as ``Google ViT'' to distinguish it from the generic ``ViT'', which encompasses other variants. In this study, we investigate 6 ViT variants, as summarized below, and provide more details in the Supplementary Material, available at https://github.com/thanhthinguyen/ecmlpkdd2024.

\subsection{Google ViT}
This ViT is a transformer encoder model, resembling the Bidirectional Encoder Representations from Transformers (BERT) \cite{kenton2019bert}, that underwent pre-training on a substantial set of images, specifically ImageNet-21k having 14 million images across 21,843 classes, with a resolution of 224$\times$224 pixels. Subsequently, fine-tuning was applied to the model using ImageNet (i.e., ImageNet Large Scale Visual Recognition Challenge - ILSVRC2012), a dataset consisting of one million images across 1,000 classes. 

\subsection{DeiT}
The \textbf{D}ata-\textbf{e}fficient \textbf{i}mage \textbf{T}ransformer (DeiT) model \cite{touvron2021training} was pre-trained and fine-tuned on the ImageNet-1k dataset, which comprises one million images at a resolution of 224$\times$224 distributed across 1,000 classes. Similar to other ViTs, this model processes images by organizing them into a sequence of linearly embedded fixed-size patches at the resolution of 16$\times$16. Through pre-training, the model obtains an internal representation of images, facilitating the extraction of features beneficial for subsequent tasks, e.g., image retrieval or classification. 

\subsection{BEiT}
The \textbf{B}idirectional \textbf{E}ncoder representation from \textbf{i}mage \textbf{T}ransformers (BEiT) model \cite{bao2022beit}, akin to BERT \cite{kenton2019bert}, is pre-trained on the ImageNet-21k dataset at a resolution of 224$\times$224 pixels. During pre-training, it focuses on predicting visual tokens derived from the encoder of OpenAI DALL-E's VQ-VAE, employing masked patches. Subsequently, the model undergoes supervised fine-tuning on the ILSVRC2012. In contrast to other ViT models, BEiT employs relative position embeddings rather than absolute position embeddings. To classify images, the mean-pooling operation is applied to the final hidden states of the patches, as opposed to using a linear layer for the final hidden state of the [CLS] token.

\subsection{DINOv2}
This is a ViT model trained using the DINOv2 method \cite{oquab2023dinov2} with DINO being a form of self-\textbf{di}stillation with \textbf{no} labels~\cite{caron2021emerging}. DINOv2 comprises a novel set of image encoders pretrained on extensive curated data without supervision. Representing an instance of self-supervised learning applied to image data, DINOv2 achieves visual features that narrow the performance gap with weakly supervised alternatives across diverse benchmarks, all without fine-tuning.

\subsection{Swin Transformer V2}
The Swin transformer constructs hierarchical feature maps by consolidating image patches in different layers \cite{liu2021swin1}. It achieves linear time complexity concerning image size by computing self-attention exclusively within each local window. To enhance scalability, Swin transformer V2 \cite{liu2022swin2} replaces the pre-norm setup in the original Swin transformer with a residual-post-norm configuration and substitutes the dot product attention with a scaled cosine attention. Furthermore, it introduces a log-spaced continuous relative position bias method to improve the model’s transferability across window resolutions.

\subsection{ViT-MAE}
This ViT model was pre-trained using the masked autoencoders (MAE) \cite{he2022masked}. Images are divided into fixed-size patches with 75\% of image patches being randomly masked out. Initially, the encoder is employed to encode these visual patches. Subsequently, a learnable (shared) mask token is introduced at the locations corresponding to the masked patches. The decoder utilizes the encoded visual patches and mask tokens as input to recreate raw pixel values at the masked positions.

\section{Knowledge Distillation for Adaptation}
\label{sec:domain_adaptation}

Hinton et al. \cite{hinton2015distilling} proposed a knowledge distillation method to transfer knowledge from a cumbersome model to a smaller student model. This approach proves valuable for compressing deep learning models and facilitating their deployment on small devices \cite{kothandaraman2021domain}. In this study, we employ Hinton et al.'s method to mitigate the issue of catastrophic forgetting during the domain adaptation process. However, it is observed that Hinton et al.'s method still experiences the forgetting problem, particularly when the target domain significantly differs from the source domain.

We propose a method (\textbf{method 1}) that allows the student to closely follow the teacher at the start of the learning process and gradually deviate as the learning progresses. In the initial phases of adjusting to a new domain, the student model may become overwhelmed with a substantial influx of new knowledge, especially in cases where the new domain diverges significantly from the source domain. Our approach aids in stabilizing the learning process of the student and results in improved performance on both the source and target domains.

\begin{figure*}[tb]
\centerline{\includegraphics[width=1.0\linewidth]{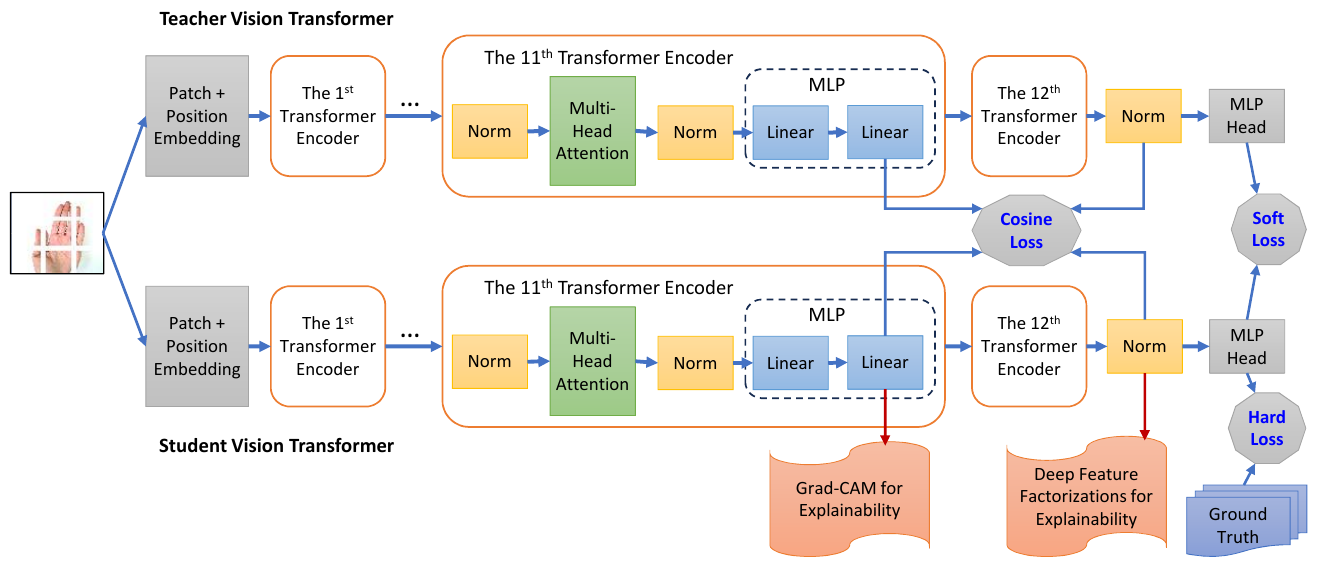}}
\caption{Domain adaptation with knowledge distilled from a teacher. The student model at the bottom is trained on data of a new domain and tries to extract knowledge from the teacher model at the top to prevent catastrophic forgetting.}
\label{fig:methodology}
\end{figure*}

We also introduce \textbf{method 2} to enable the student to glean insights from the internal states of the teacher model. Given that the teacher model harbors a superior hidden representation, which is difficult for the student to achieve unaided, we deliberately guide the student to replicate not only the teacher's output layers but also its internal states. Our adaptive distillation methods are summarized in Fig. \ref{fig:methodology} and detailed in the following subsections.  

\subsection{Method 1: Adaptive distillation}
A deep learning model for a classification task normally receives an input variable $x$ and generates an output logit $z_i$ where $i \in [1, C]$ with $C$ being the number of classes. To identify the class label based on the pseudo-probabilities $y_i$, this unnormalized logit is passed through a softmax function so that:
\begin{equation}
y_i(x|T) = \frac{e^{z_i/T}}{\sum_{j=1}^{C} e^{z_j/T}}
\label{eq:1}
\end{equation}
where $T$ is the temperature parameter. A standard softmax function uses a temperature of 1, but a higher temperature will create a softer distribution over the classes where classes with small probabilities also have some effect. A student (distilled model) can extract knowledge from the teacher model (knowledge distillation) when training on a transfer dataset using a cross entropy loss (soft loss) between the student output logit and the teacher output logit on the same training sample $x$ using the same high value of temperature $T$. Let us denote the pseudo-probabilities created by the student and teacher models, which can be computed using Eq. \ref{eq:1}, as $y(x|T)$ and $\hat{y}(x|T)$, respectively. The soft loss $\mathcal{L}_s$ is then defined as:
\begin{equation}
\mathcal{L}_s(\hat{y}, y) = -\sum_{i=1}^C \hat{y_i}(x|T)\log y_i(x|T)
\label{eq:soft}
\end{equation}
A hard loss $\mathcal{L}_h$ between the student output logit $z$ and the ground truth label, with $g$ being the index of the label, is specified as:
\begin{equation}
\mathcal{L}_h(z,g) = -w_{g}\log \frac{\exp(z_{g})}{\sum_{i=1}^C \exp(z_{i})}
\label{eq:hard}
\end{equation}
where $w_g$ is a weight assigned to the class $g$. We set equal weights to classes in our experiments so that $w_g=1$, $\forall g \in [1, C]$ in this study.

A weighted average of these two losses (i.e., $\mathcal{L}_s$ and $\mathcal{L}_h$) can be used for fine-tuning the student model using the transfer set. A significantly higher weight for the soft loss, i.e., multiply it by $T^2$, is used because the magnitudes of the gradients generated by the soft targets scale by a factor of $1/T^2$. The overall loss is therefore:
\begin{equation}
\mathcal{L} = T^2\mathcal{L}_s + \mathcal{L}_h
\label{eq:4}
\end{equation}
We propose a distillation process to stabilize the learning process of the student model by the following overall loss function:
\begin{equation}
\mathcal{L} = \sqrt{\frac{E}{e}}T^2\mathcal{L}_s + \mathcal{L}_h
\label{eq:5}
\end{equation}
where $E$ is the total number of training epochs and $e$ is a variable $e \in [1, E]$ representing the epoch being learned. At the beginning of the learning process, $e$ is equal to 1 so that the model allocates greater focus (more weight) to imitating the teacher behaviours via the soft loss $\mathcal{L}_s$. The model will increasingly acquire new knowledge when $e$ increases from 1 to its maximum value $E$. At the final epoch, i.e., $e=E$, the loss defined in Eq. \ref{eq:5} is equivalent to that in Eq. \ref{eq:4}.

\subsection{Method 2: Imitate the teacher's internals}
\label{subsec:method2}
A deep learning model designed for a computer vision task typically harbors valuable internal information regarding the features and characteristics of the data, contributing to the interpretation of model outputs. While emulating the teacher's behaviors based on its generated outputs is valuable, there is potential merit in also replicating the internal representations/states of the teacher. Using the deep feature factorization (DFF) and Grad-CAM tools, we evaluate the explainability of different components of the ViT model. The DFF method involves decomposing the activations from the model into distinct concepts (e.g., class labels) using non-negative matrix factorization. Subsequently, for each pixel, the method calculates its correlation with each of these labels. This technique is adept at pinpointing analogous semantic concepts inside an individual image or a collection of images \cite{collins2018deep}. On the other hand, Grad-CAM is employed to visualize the regions of an input image that contribute the most to the predictions made by a neural network. The Grad-CAM heatmap provides a visual representation of the areas that the network focused on during its decision-making process \cite{selvaraju2020grad}. DFF and Grad-CAM were originally proposed to deal with CNN models. In this study, we use these two methods to acquire a more profound insight into the learned features of the ViT models. 

A Google ViT model has 12 layers of transformer encoders. Through numerous experiments examining various components at different layers of the ViT, we could verify the usefulness of the final linear component of the 11th transformer encoder layer and the layer normalization (or Norm in Fig.~\ref{fig:methodology}) component of ViT for explainability purpose as suggested in \cite{jacobgilpytorchcam}. We therefore propose an approach to enable the student to mimic these two components of the teacher. 

Let us denote the bias and weight parameters of the above-mentioned Norm as $b_{n}$ and $w_{n}$, the bias and weight parameters of the final linear component of the 11th layer (i.e., the second last layer) of the ViT as $b_{o}$ and $W_{o}$, respectively. Let $H$ and $I$ be the hidden size and intermediate size of the ViT configuration, $b_{n}$, $w_{n}$, and $b_{o}$ are 1-dimension arrays at length of $H$ while $W_{o}$ is a 2-dimension matrix with a size of $H\times I$. We transpose and get a mean of $W_{o}$ before stacking it with the remaining parameters arrays. The resulting matrix is therefore a stack of $b_{n}$, $w_{n}$, $b_{o}$, and $\overline{{W_{o}^T}}$, and has a size of 4$\times H$. The same procedure is applied for both student model and the teacher model, resulting in two embedding matrices $W_s$ and $W_t$ for the student and teacher, respectively. The cosine embedding loss is used to measure whether the two input matrices are similar or dissimilar, using the cosine similarity, defined as below:
\begin{equation}
\mathcal{L}_c(W, y) =
    \begin{cases}
        1 - \cos(W_s, W_t), & \text{if } y = 1 \\
        \max(0, \cos(W_s, W_t)), & \text{if } y = -1
    \end{cases}
    \label{eq:cosineloss}
\end{equation}
where $y$ is the target, which is set equal to 1 in all experiments in this study. To enable an adaptive distillation for domain adaptation, an overall loss function is specified as:
\begin{equation}
\mathcal{L} = \sqrt{\frac{E}{e}} \left( T^2\mathcal{L}_s+\mathcal{L}_c \right) + \mathcal{L}_h
\label{eq:9}
\end{equation}
where $\mathcal{L}_s$, $\mathcal{L}_h$, and $\mathcal{L}_c$ are defined in Eqs. \ref{eq:soft}, \ref{eq:hard}, and \ref{eq:cosineloss}, respectively. This overall loss function is used to fine-tune the student model on the transfer set. In this study, we use the Google ViT as the student model. This ViT has been trained on the source domain data. To adapt this model to data of a different domain, we experiment two scenarios for the teacher. The first scenario for the teacher uses a soft voting ensemble of 6 ViT variants (each described in section \ref{sec_vit_models}) that all have been trained on the source domain data. The average of pseudo-probabilities across individual methods is used to calculate the soft loss in Eq. \ref{eq:soft} in this scenario. The second scenario uses a copy of the student itself as the teacher. In both scenarios, the teacher has a great knowledge of the source domain data, similar to the student itself before adapting to the new domain. With the knowledge distilled from the teacher, we enable the student to avoid forgetting the source domain knowledge while learning on data of the new domain.

\section{Hand Images Datasets}
\label{sec_data}
\subsection{IIT Delhi Dataset}
The first dataset used in our experiments is the IIT Delhi touchless palm print image dataset version 1.0 \cite{iitd_dataset,kumar2008incorporating}. This dataset comprises images captured from hands of students and staff at IIT Delhi, India by a digital CMOS camera during January 2006 - July 2007. Left and right hand colour images are obtained from 230 subjects through a touchless imaging setup at a resolution of 1600$\times$1200. Subjects have ages from 14 to 56 years old and each contributed minimum 5 palmar hand images from both left and right hands. 

\subsection{11k Hands Dataset}
The second dataset is the 11k hands dataset. This dataset was introduced in \cite{afifi201911k}, which includes hand images from 190 subjects with both left and right hands, and both palmar and dorsal sides at a resolution of 1600$\times$1200. This is the most recent dataset in this domain, consisting of 11,076 colour hand images of various subjects with ages in the range from 18 to 75. This dataset was published with detailed metadata, which records not only the sides of the hands (e.g., dorsal or palmar, left or right) but also information about whether hand images contain accessories, irregularities, or nail polish.

\section{Experimental Results}
\label{sec_exp_res}

We adopt the classification accuracy as the performance evaluation metric for comparisons between ViT models and traditional methods. Each fine-tuning process is conducted through 50 epochs, using an Adam optimizer at a learning rate of $2 \times 10^{-5}$. Images are randomly cropped and horizontally flipped for data augmentation during training with a batch size of 32.

\subsection{ViT models vs traditional methods}
\subsubsection{IIT Delhi dataset}
To benchmark our work with existing methods, e.g., \cite{afifi201911k,charfi2017bimodal,bera2017finger}, we perform three experiments corresponding to classifying hand images of 100, 137 and 230 subjects. For each subject, we select randomly four images for the training set and one image for the testing set. For a objective evaluation of the ViT models, the experiment is repeated 10 times for each model with the subjects being randomly selected each time. Results shown in Table \ref{table:iitd} are the averaged accuracy and standard deviation across 10 times.

\begin{table}[htbp]
\centering
\begin{scriptsize}
\caption{Hand image classification results using the IIT Delhi dataset. Traditional methods (whose results extracted from \cite{afifi201911k}) use hand-crafted features and/or CNN-based deep features while the ViT models divide images into patches.}
\begin{tabular}{|l|l|c|c|c|}
\hline
 \textbf{Methods} & \textbf{Features} & \textbf{S=100} & \textbf{S=137} & \textbf{S=230} \\ 
\hline
\rowcolor{Gray}
 Kumar and Shekhar \cite{kumar2010palmprint} & Gabor & - & - & 0.950 \\
\hline
\rowcolor{Gray}
 Kumar and Shekhar \cite{kumar2010palmprint} & Radon transform & - & - & 0.928 \\ 
\hline
\rowcolor{Gray}
 Charfi et al. \cite{charfi2014novel} & SIFT & - & - & 0.940 \\ 
\hline
\rowcolor{Gray}
 Bera et al. \cite{bera2017finger} & Finger contour profile & - & 0.978 & 0.952 \\ 
\hline
\rowcolor{Gray}
 Charfi et al. \cite{charfi2017bimodal} & Standard SIFT & 0.852 & - & 0.803 \\ 
\hline
\rowcolor{Gray}
 Charfi et al. \cite{charfi2017bimodal} & SIFT+SR & 0.962 & - & - \\ 
\hline
\rowcolor{Gray}
 Afifi \cite{afifi201911k} & CNN-features & 0.891 & 0.912 & 0.900 \\ 
\hline
\rowcolor{Gray}
 Afifi \cite{afifi201911k} & CNN-features+LBP & 0.940 & 0.964 & 0.948 \\ 
\hline
Google ViT \cite{dosovitskiy2021an} & Image patches & 0.994 $\pm$ 0.008 & \textbf{0.998} $\pm$ 0.003 & 0.996 $\pm$ 0.004 \\
\hline
\rowcolor{LightCyan}
DeiT \cite {touvron2021training} & Image patches & \textbf{0.997} $\pm$ 0.005 & \textbf{0.998} $\pm$ 0.005 & 0.998 $\pm$ 0.002 \\
\hline
\rowcolor{LightCyan}
BEiT \cite{bao2022beit} & Image patches & \textbf{0.997} $\pm$ 0.005 & 0.995 $\pm$ 0.007 & \textbf{0.999} $\pm$ 0.002 \\
\hline
DINOv2 \cite{oquab2023dinov2} & Image patches & 0.982 $\pm$ 0.018 & 0.976 $\pm$ 0.018 & 0.912 $\pm$ 0.107 \\
\hline
\rowcolor{LightCyan}
Swin Transformer V2 \cite{liu2022swin2} & Image patches & 0.995 $\pm$ 0.009 & 0.996 $\pm$ 0.006 & \textbf{0.999} $\pm$ 0.002 \\ 
\hline
ViT-MAE \cite{he2022masked} & Image patches & 0.986 $\pm$ 0.010 & 0.948 $\pm$ 0.079 & 0.987 $\pm$ 0.006\\ 
\hline
\end{tabular}
\label{table:iitd}
\end{scriptsize}
\end{table}

Table \ref{table:iitd} shows that most ViT models outperform traditional methods, with DeiT, BEiT and Swin Transformer V2 being the most accurate and stable methods. Specifically, in the 100-subject experiment, the best existing accuracy is 0.940 from the work in \cite{afifi201911k}, whilst ViT models obtain an accuracy of at least 0.982, i.e., the result of DINOv2. The best ViT models in this experiment are DeiT and BEiT with both achieving an accuracy of 0.997. 

\subsubsection{11k hands dataset}

We compare ViT models with the existing method \cite{afifi201911k} using the 11k hands dataset. Six experiments were conducted in \cite{afifi201911k}, including dorsal-80, palm-80, dorsal-100, palm-100, dorsal-120, and palm-120, where dorsal-80 means the task is to classify 80 subjects based on their hand dorsal-side images, whilst palm-80 is to classify 80 subjects using their hand palm-side images. For comparison purpose, we use exactly the same training and testing sets created in \cite{afifi201911k}. Specifically, a training set contains 10 hand images, whilst a testing set consists of four images, and each experiment is rerun 10 times with subjects and images being chosen randomly each time. 

\begin{table}[htbp]
\centering
\begin{scriptsize}
\caption{Hand image classification results on the 11k hands dataset. On average, the ViTs exhibit better prediction accuracy when presented with images of the dorsal side as opposed to those capturing the palm side.}
\begin{tabular}{|l|c|c|c|c|c|c|}
\hline
 \textbf{Methods} & \textbf{dorsal-80} & \textbf{palm-80} & \textbf{dorsal-100} & \textbf{palm-100} & \textbf{dorsal-120} & \textbf{palm-120} \\ 
\hline
\rowcolor{Gray}
 CNN/SVM \cite{afifi201911k} & 0.96 $\pm$ 0.01 & 0.95 $\pm$ 0.02 & 0.96 $\pm$ 0.01 & 0.93 $\pm$ 0.02 & 0.96 $\pm$ 0.01 & 0.93 $\pm$ 0.01 \\
\hline
\rowcolor{Gray}
 CNN+LBP/SVM \cite{afifi201911k} & 0.96 $\pm$ 0.02 &  0.96 $\pm$ 0.01 & 0.97 $\pm$ 0.01 & 0.95 $\pm$ 0.01 & 0.97 $\pm$ 0.01 & 0.96 $\pm$ 0.01\\ 
\hline
\rowcolor{LightCyan}
Google ViT \cite{dosovitskiy2021an} & 0.99 $\pm$ 0.01 & \textbf{0.99} $\pm$ 0.01 & 0.99 $\pm$ 0.01 & \textbf{0.99} $\pm$ 0.01 & 0.99 $\pm$ 0.00 & \textbf{0.99} $\pm$ 0.01\\
\hline
DeiT \cite {touvron2021training} & 0.99 $\pm$ 0.01 & 0.99 $\pm$ 0.01 & 0.99 $\pm$ 0.00 & 0.98 $\pm$ 0.01 & 0.99 $\pm$ 0.00 & 0.98 $\pm$ 0.01\\
\hline
\rowcolor{LightCyan}
BEiT \cite{bao2022beit} & \textbf{0.99} $\pm$ 0.01 & \textbf{0.99} $\pm$ 0.01 & \textbf{0.99} $\pm$ 0.01 & \textbf{0.99} $\pm$ 0.01 & \textbf{1.00} $\pm$ 0.00 & \textbf{0.99} $\pm$ 0.01\\
\hline
DINOv2 \cite{oquab2023dinov2} & 0.98 $\pm$ 0.02 & 0.95 $\pm$ 0.03 & 0.98 $\pm$ 0.03 & 0.97 $\pm$ 0.03 & 0.97 $\pm$ 0.03 & 0.95 $\pm$ 0.03 \\
\hline
\rowcolor{LightCyan}
Swin Trans. V2 \cite{liu2022swin2} & \textbf{0.99} $\pm$ 0.01 & 0.98 $\pm$ 0.01 & \textbf{1.00} $\pm$ 0.00 & 0.99 $\pm$ 0.01 & \textbf{1.00} $\pm$ 0.00 & 0.98 $\pm$ 0.01\\ 
\hline
ViT-MAE \cite{he2022masked} & 0.97 $\pm$ 0.02 & 0.97 $\pm$ 0.01 & 0.98 $\pm$ 0.01 & 0.96 $\pm$ 0.01 & 0.98 $\pm$ 0.00 & 0.97 $\pm$ 0.01\\ 
\hline
\end{tabular}
\label{table:11k}
\end{scriptsize}
\end{table}

Table \ref{table:11k} shows the results obtained by ViT models in comparisons with two approaches proposed in \cite{afifi201911k}, in which the first approach uses only CNN features and a SVM classifier whilst the second approach also uses a SVM classifier but the feature set is a combination between CNN features and LBP features. The results demonstrate that a majority of ViT models significantly dominate the conventional methods in \cite{afifi201911k}. The most superior methods are the Google ViT, BEiT and Swin Transformer V2.  BEiT produces the best performance in 4 of 6 experiments, while the Google ViT and Swin Transformer V2 share the second best position, showing the superior performance in 3 out of 6 experiments. 

\subsection{ViT explainability results}

\begin{figure}
\centering
\begin{subfigure}{0.68\linewidth}        
\centerline{\includegraphics[width=1.0\linewidth]{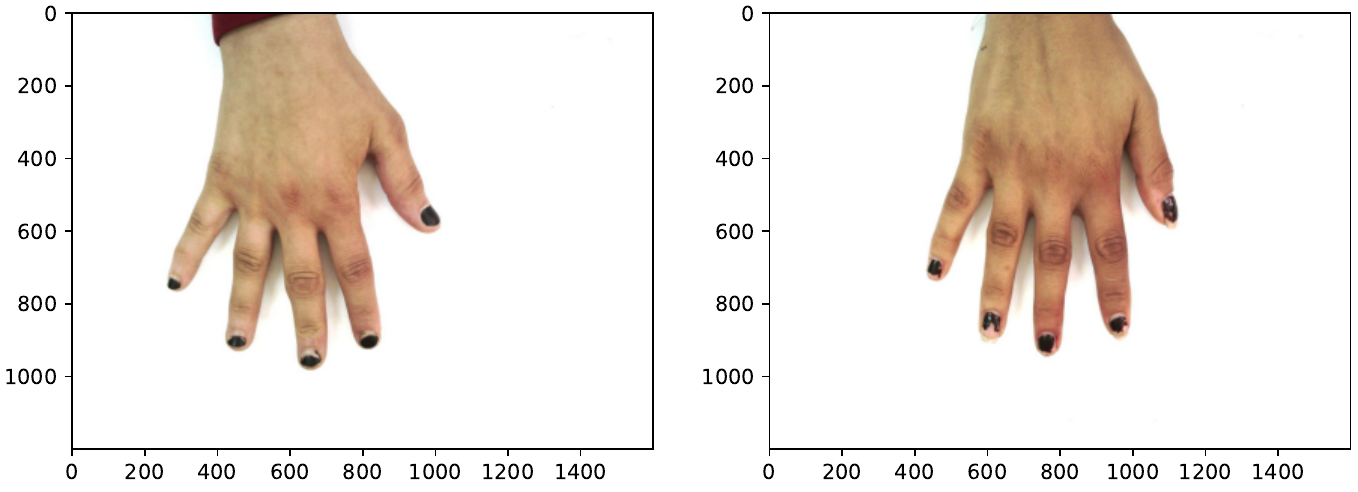}}        
\caption{Dorsal image of subject `0001046' (left), and subject `0001573' (right)}\label{fig:image21}
\end{subfigure} 
\begin{subfigure}{0.68\linewidth}
\centerline{\includegraphics[width=1.0\linewidth]{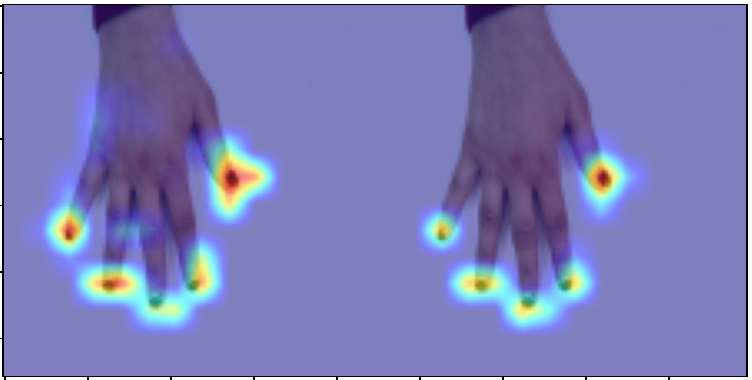}}          
\caption{A Grad-CAM explanation shows how the ViT model identifies subject `0001046' based on the nail polish on their fingers.}
\label{fig:image22}
\end{subfigure}
\begin{subfigure}{0.68\linewidth}
\centerline{\includegraphics[width=1.0\linewidth]{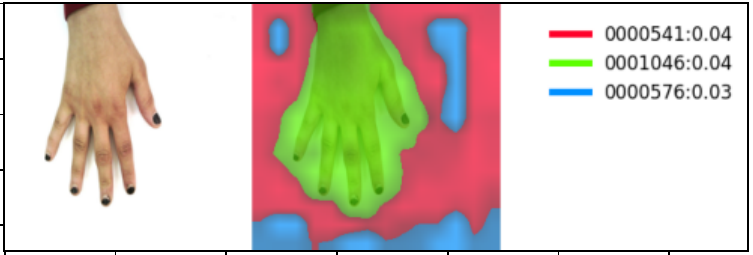}}          
\caption{Explainability with the deep feature factorization visualization.}
\label{fig:image23}
\end{subfigure}
\caption{Explainability of a ViT using the PyTorch library for CAM methods.}
\label{fig:rightdorsal_explain}
\end{figure}

An illustration of the explainability results based on the DFF and Grad-CAM methods is shown in Fig. ~\ref{fig:rightdorsal_explain}. The PyTorch library for CAM methods \cite{jacobgilpytorchcam} is used to investigate the internal representations of different components of the Google ViT model. Fig. \ref{fig:image21} presents raw dorsal right hand images of subjects ``0001046'' (left image), and subject ``0001573'' (right image) in the 11k hands dataset.

Fig. \ref{fig:image22} illustrates the internal representations of the ViT when attempting to predict the subject of an input image, which is the image on the left of Fig. \ref{fig:image21}. This internal state of the ViT is extracted from the parameters (after fine-tuning) of the second linear component of the 11th encoder layer of the ViT (i.e., the linear component contributes to the cosine loss in Fig.~\ref{fig:methodology}). This component helps to explain that the ViT predicts subject ``0001046'' (which is correct) based on nail polish on the fingers. Given the input image, the left image of Fig. \ref{fig:image22} illustrates the ViT's interpretation of the subject ``0001046'' while the right image of Fig. \ref{fig:image22} depicts its perspective on the subject ``0001573''. Both left and right images of Fig. \ref{fig:image22} indicate nail polish as the features for recognizing the subjects ``0001046'' and ``0001573'', which is correct because both subjects have nail polish on their fingers (as shown in Fig. \ref{fig:image21}). However, the ViT model might encounter confusion in distinguishing between these two subjects when relying solely on information from the mentioned linear component.

How does the ViT predict that the input image belongs to subject ``0001046'' rather than subject ``0001573''? We found that this can be explained by another component, which is the Norm of the ViT (the Norm contributes to the cosine loss in Fig. \ref{fig:methodology}). Based on parameters of this component, the DFF method helps to explain (as in Fig.~\ref{fig:image23}) that the entire region of the hand in the image (with green color) is predicted to be of subject ``0001046''. The shape of the dorsal hand, consisting of the length and width of fingers, could have been helpful for the ViT (via the Norm component) to differentiate between subjects ``0001046'' and ``0001573''. Given the usefulness of the second linear component of the 11th encoder layer and the Norm component of the ViT, we utilize them to develop the adaptive knowledge distillation method 2 to imitate the teacher's internals, with results presented in subsection \ref{sec:adaptation}.

\subsection{ViT models for transfer inference}
\label{subsection:inference}

We conduct several experiments to demonstrate the accuracy of the ViT models when they are transferred to perform inference on data of a different domain. 

As the IIT Delhi dataset has only palm-side images, we choose randomly four \textit{left-hand} images for each subject to create a source dataset. We aim to transfer the trained models to predict labels on a target dataset, which comprises four randomly selected \textit{right-hand} images for each subject.

The 11k hands dataset has both palm- and dorsal-side images of left and right hands, and we therefore design experiments to transfer models not only from left to right side, but also from palm to dorsal side, and vice versa. Specifically, we train models on \textit{left-palm} images and transfer them to predict labels for \textit{right-palm} and \textit{left-dorsal} images. Similarly, we train models on \textit{right-dorsal} images and predict labels for \textit{left-dorsal} and \textit{right-palm} images. We create a source dataset by selecting randomly four images for each subject. Each target dataset also consists of four randomly selected images for each subject.

\begin{table}[htbp]
\centering
\begin{scriptsize}
\caption{Transfer inference results for one experimental scenario for the IIT Delhi dataset and four scenarios for the 11k hands dataset. Transferring from left to right or right to left obtains better accuracy than transferring from palm to dorsal or dorsal to palm.}
\begin{tabular}{|l|c|c|c|c|c|}
\hline
 \textbf{Methods} & \textbf{\makecell{IIT Delhi\\Left$\rightarrow$Right}} & \textbf{\makecell{11k (Palm)\\Left$\rightarrow$Right}} & \textbf{\makecell{11k (Dorsal)\\Right$\rightarrow$Left}} & \textbf{\makecell{11k (Left)\\Palm$\rightarrow$Dorsal}} & \textbf{\makecell{11k (Right)\\Dorsal$\rightarrow$Palm}} \\ 
\hline
Google ViT \cite{dosovitskiy2021an} & 1.00 $\rightarrow$ 0.81 & 0.99 $\rightarrow$ 0.49 & 0.99 $\rightarrow$ 0.55 & 0.99 $\rightarrow$ 0.04 & 0.99 $\rightarrow$ 0.08\\
\hline
DeiT \cite {touvron2021training} & 1.00 $\rightarrow$ 0.77 & 0.98 $\rightarrow$ 0.41 & 0.99 $\rightarrow$ 0.55 & 0.98 $\rightarrow$ 0.14 & 0.99 $\rightarrow$ 0.09\\
\hline
BEiT \cite{bao2022beit} & 1.00 $\rightarrow$ 0.83 & 0.96 $\rightarrow$ 0.29 & 0.99 $\rightarrow$ 0.51 & 0.96 $\rightarrow$ 0.05 & 0.99 $\rightarrow$ 0.10 \\
\hline
DINOv2 \cite{oquab2023dinov2} & 0.99 $\rightarrow$ 0.60 & 0.89 $\rightarrow$ 0.20 & 0.99 $\rightarrow$ 0.49 & 0.89 $\rightarrow$ 0.05 & 0.99 $\rightarrow$ 0.04 \\
\hline
Swin Trans. V2 \cite{liu2022swin2} & 1.00 $\rightarrow$ 0.80 & 0.98 $\rightarrow$ 0.38 & 0.99 $\rightarrow$ 0.57 & 0.98 $\rightarrow$ 0.12 & 0.99 $\rightarrow$ 0.08 \\ 
\hline
ViT-MAE \cite{he2022masked} & 1.00 $\rightarrow$ 0.77 & 0.93 $\rightarrow$ 0.27 & 0.98 $\rightarrow$ 0.45 & 0.93 $\rightarrow$ 0.09 & 0.98 $\rightarrow$ 0.07\\ 
\hline
\rowcolor{LightCyan}
\textbf{Soft Voting} & 1.00 $\rightarrow$ 0.90 & 0.98 $\rightarrow$ 0.40 & 1.00 $\rightarrow$ 0.67 & 0.98 $\rightarrow$ 0.08 & 1.00 $\rightarrow$ 0.08\\ 
\hline
\end{tabular}
\label{table:transfer}
\end{scriptsize}
\end{table}

Results of the transfer inference experiments are reported in Table \ref{table:transfer}. In each column, the left values show inference results of ViTs on the source domain, while the right values show those on the target domain. The soft voting ensemble method demonstrates enhanced accuracy in several experiments, though not consistently across all of them. Transferring from left to right in the IIT Delhi dataset obtains the best results with accuracy on the target domain around 0.6 to 0.8. Transferring from left to right or right to left in the 11k dataset achieves lower accuracy, about 0.2 to 0.6. It demonstrates that images in the 11k dataset are more challenging to learn from compared with those in the IIT Delhi dataset. Notably, the transfer results from palm to dorsal or dorsal to palm are extremely poor with the accuracy on the target domain approximately 0.1. Applying our knowledge distillation methods in domain adaptation can contribute to improving this subpar inference performance (results are shown in subsection \ref{sec:adaptation}). 

\subsection{ViT domain adaptation}
\label{sec:adaptation}

This subsection presents results of our adaptive distillation methods (methods 1 and 2) for domain adaptation in hand image classification in comparison with the baseline, i.e., the Hinton et al.'s method \cite{hinton2015distilling}. The temperature $T$ equal to 2 is selected to compute the soft loss for all experiments in this subsection. We conducted 5 experimental scenarios corresponding to those in subsection \ref{subsection:inference}. In each scenario, we check the accuracy of the student before learning to adapt to a new domain. We then conduct the adaptation by training the model on the target domain without any distillation, aiming to assess the extent to which the model forgets knowledge from the source domain. Experiments for comparisons between our methods 1 and 2 with the Hinton et al.'s method are followed with the teacher being either the soft voting ensemble or the student's prior copy. The ensemble teacher does not have a specific internal representation so that our method 2 is not conducted in that case.

In the first three scenarios (i.e., left-palm to right-palm on the IIT Delhi dataset, left-palm to right-palm and righ-dorsal to left-dorsal on the 11k hands dataset), the student model retains significant knowledge of the source domain after training on the target domain even in the absence of distillation. After training on the target domain, the accuracy of the student on the source and target domains in these three scenarios are around 0.95 to 1.0. Catastrophic forgetting is minimal in these scenarios because the two domains (left and right) are closely related. The distillation methods provide improvement, although their impact is not substantial in these scenarios. Detailed results for these three scenarios and the left-palm to left-dorsal scenario using the 11k hands dataset are presented in Supplementary Material, at https://github.com/thanhthinguyen/ecmlpkdd2024. 

\begin{table}
  \centering
  \begin{footnotesize}
    \caption{Results on adapting from right \textit{dorsal} domain to right \textit{palm} domain in the 11k hands image dataset.}
  \begin{tabular}{@{}lcc@{}}
    \toprule
    Accuracy of & \makecell{Source Domain\\Right Dorsal} & \makecell{Target Domain\\Right Palm} \\
    \midrule
    \makecell[l]{Student before adapting} & 0.992 & 0.075 \\
    \makecell[l]{Ensemble teacher} & 0.996 & 0.083 \\
    \hline
    Student after adapting - no distil & 0.472 & 0.975\\
    \rowcolor{Gray} 
    \multicolumn{3}{l}{Student after adapting with knowledge distilled from ...} \\    
    the ensemble teacher (Hinton et al.) & 0.681 & 0.960 \\
    the ensemble teacher (method 1) & 0.731 & 0.960 \\
    \hline
    prior copy as teacher (Hinton et al.) & 0.767 & 0.964 \\
    prior copy as teacher (method 1) & 0.888 & 0.968 \\
    prior copy as teacher (method 2) & 0.893 & 0.971 \\
    \bottomrule
  \end{tabular}
  \label{tab:adapt_rightdorsalpalm}
  \end{footnotesize}
\end{table}

Results for the right-dorsal to right-palm adaptation are presented in Table~\ref{tab:adapt_rightdorsalpalm}. When training on the target domain without distillation, the student model's accuracy on the source domain drops to 0.472. It indicates the occurrence of catastrophic forgetting, a phenomenon that is understandable given the considerable dissimilarity between the two domains (palm and dorsal). We observed that distillation methods result in significant improvements in this scenario (and also left-palm to left-dorsal scenario presented in the Supplementary Material). Particularly, methods 1 and 2 achieve comparable or superior accuracy on both the source and target domains when compared to Hinton et al.'s method. Our methods contribute to a noteworthy improvement on the source domain by mitigating the catastrophic forgetting problem. 

The student extracts knowledge from the ensemble teacher via their voting results. Even though these voting results are stable and sometimes better than the individual method, distilling knowledge from the ensemble is less effective than using a prior copy of the student as the teacher. This is because it is more challenging to interpret the internal states of the ensemble teacher than the individual teacher based on their outputs. 
This helps to elucidate why method 2 outperforms method 1 in the majority of experiments, as method 2 compels the student to replicate the internals of the teacher.

\section{Conclusion and Future Work}
We have fine-tuned 6 ViT models for classification of hand images. A stable and superior performance of ViT models has been demonstrated compared with existing methods. We show the usefulness of two components of ViT for the explainability purpose. Results also demonstrate the effectiveness of our distillation methods when training a ViT model on a new domain. Our methods help to obtain high classification performance on both source and target domain even the model has no access to the source domain data. A further work can be to develop our methods on other types of data such as fingerprints and iris recognition data for biometric applications.

\begin{credits}
\subsubsection{\ackname} Portions of the work tested on the IITD Touchless Palmprint Database version 1.0, available in \cite{iitd_dataset}, and the 11k hands images dataset in \cite{afifi201911k}. This research/project was undertaken with the assistance of resources and services from the National Computational Infrastructure (NCI), which is supported by the Australian Government. 
\subsubsection{\ethicname} While our hand image classification algorithms offer benefits in applications such as access control, identity verification, and authentication systems, their real-world implementation must be guided by ethical principles. There are risks associated with the potential usage of our methods for biometric identification. They include lack of informed consent, discrimination and bias, privacy concerns, misuse and abuse, among other risks.
\begin{enumerate}[label=\arabic*)]
\item Lack of Informed Consent: Individuals may not fully understand the implications of handing over their hand images for biometric identification, particularly in the context of police surveillance. Hence, transparent information about the purpose, risks, and implications of such biometric identification needs to be provided to the involved individuals.
\item Discrimination and Bias: Machine learning-based biometric systems, including those using hand images, may exhibit bias or inaccuracies, leading to disproportionate targeting or misidentification of certain demographics, such as people of colour or individuals with disabilities. Biometric systems therefore need to be regularly evaluated and audited. Diverse representation is needed in the development and testing of those systems.
\item Privacy Concerns: Hand images can be used to identify individuals. Collecting and storing such data for surveillance purposes raises privacy concerns. Strict regulations need to be implemented for governing the collection, storage, and use of hand images for biometric identification.
\item Misuse and Abuse: Biometric data, including hand images, can be misused or abused if it falls into the wrong hands. Robust security measures need to be employed to protect biometric databases from unauthorized access or breaches. Employing encryption, access controls, and conducting routine security audits can mitigate the chances of misuse.
\end{enumerate}

By implementing robust regulations, security measures, and transparency mechanisms, it may be possible to harness the benefits of using our hand image classification methods while minimizing its ethical drawbacks in police surveillance contexts.

\end{credits}
%
%
%


 

\bibliographystyle{splncs04}
\bibliography{main}

%




\end{document}